\documentclass{article}
\pdfoutput=1 
\usepackage[english]{babel}
\usepackage[T1]{fontenc}
\usepackage[utf8x]{inputenc}

\usepackage{amsmath,amsfonts,amssymb}
\usepackage{hyperref}
\usepackage{graphicx}
\usepackage{subcaption}
\usepackage{a4wide}

\title{Metasensor: a proposal for sensor evolution in robotics}
\author{Michele Braccini$^1$}
\date{%
    $^1$Department of Computer Science and Engineering, Universit{\`a} di Bologna,
Campus~of~Cesena, I-47521 Cesena, Italy;\\%
}

\begin{document}

\maketitle

\begin{abstract}
Sensors play a fundamental role in achieving the complex behaviors typically found in biological organisms.
However, their potential role in the design of artificial agents is often overlooked.
This often results in the design of robots that are poorly adapted to the environment, compared to their biological counterparts.
This paper proposes a formalization of a novel architectural component, called a metasensor, which enables a process of sensor evolution reminiscent of what occurs in living organisms.
Even in online scenarios, the metasensor layer searches for the optimal interpretation of its input signals and then feeds them to the robotic agent to accomplish the assigned task.
\end{abstract}

\section{Introduction}
Since its introduction, cybernetics has been concerned with investigating the mechanisms that enable the control and adaptation of biological and artificial organisms.
On the other hand, biosemiotics considers symbols and signs as essential elements in the economies of living beings.
Their use allows them to communicate, adapt and evolve within their biological and ecological environments.

The topic of sensory system evolution~\footnote{By ``evolution'' we mean any procedure, either offline or online, that may allow sensors to be modified so as to make the robot better fitted to the environment in which it operates: so, in the following---even when not explicitly stated---it also includes processes of development, adaptation and learning that may occur during the robot's lifetime.} naturally fits between the folds of the space traced by both the aforementioned disciplines.
Indeed, sensors are part of control in that they are part of---along with controller, actuators and environment---the causal closed-loop that composes the feedback mechanism required to achieve the desired behavior of the robot (cybernetic dimension).
At the same time, being the interface between the external world and the internal system, they process---providing a first implicit interpretation---the signs coming from the external world (semiotic dimension).
In a truly open-ended scenario~\cite{kauffman2019world,kauffman2021theorem,cortes2023tap,pattee2019evolved}, signs and their internal interpretations evolve as a consequence of the evolution of the nature and quality of the sensors themselves, changing the behavior of the robotic agent accordingly and vice versa.

So, although sensory apparatus plays a fundamental role in the evolution and adaptation of biological organisms and there is evidence to suggest that they can play an equally important role in the design of robotic behavior~\cite{balakrishnan1996onsensor,olsson2005sensor,roubieu2014biomimetic}, their potential role, for purely practical reasons, is often overlooked in favor of control software design~\cite{pfeifer2014cognition,pfeifer2001understanding}.
The common practice in robotics is to adjust the robot controller to reduce the gap between actual and desired behavior: sensors appear as designer-imposed properties of the robot---expressed in terms of sensors number, type, and placement---and not degrees of freedom on which the robot can act to achieve its goal.
Although there are approaches in which co-evolution of robot controller and morphology is present~\cite{auerbach2011evolving,bongard2003evolving,bert2010automatically,jin2011morphogenetic,pagliuca2022dynamic,parker2007coevolution}, the question of sensor evolution often boils down to a particular morphological composition of predetermined sensor building blocks, a specific choice among the combinatorial possibilities of sensors offered by the designer's sensor library.

Excluding sensor modifications from the design of robots leads to a reduction in the search space of candidate solutions, which ultimately translates into robots less fitted to their environment and less robust in maintaining their behavior~\cite{thompson1999explorations} than biological organisms.
Their exclusion also has implications for the degree and evolution~\cite{thompson1997hardware} of the fault tolerance property of a system.
Sensor failures may indeed occur in robot operating in real time, controller adaptation mechanisms alone might not be sufficient to cope with all potential contingencies.
In this scenario, the online---during robot lifetime---evolution of sensors could be a game changer.

So far, since Gordon Pask's remarkable but unique attempts to evolve real physical sensors in 1958~\cite{cariani1993evolve,pask1958growth,pask1960natural,pask1958physical} and the ``intrinsic hardware evolution'' described by Adrian Thompson~\cite{thompson1996uncostrained,thompson1997evolved,thompson1997hardware,thompson1999explorations}, little has been done in this direction.
The primary reason for the paucity of works dealing with the artificial evolution of sensors is due to the impossibility of having real changes in the physical substrate representing the sensory apparatus.
The examples of artificial sensor evolution that can be found in the literature are limited---both in number and relevance---and only scratch the surface of the topic of interest.

Olsson et al.~\cite{olsson2005sensor} proposed an effective but limited sensory adaptation method.
Indeed, since it is not possible to expand the sensory capabilities provided by the system designer or changing their physical structures, the method tries to optimize its use.
More precisely, through Shannon entropy maximization and adaptive binning, the robot continuously adapts its internal transfer function to the structure of the input distribution in order to produce optimal and compressed encoding of sensory information.

In~\cite{balakrishnan1996onsensor}, the authors instead presented an offline design method for sensors involving the employment of a genetic algorithm with operators allowed to operate not only on the structure of the control software architecture but also on the positions and ranges of the sensors, all of the same type.
Their work suggests an advantage in combined optimization of controllers and sensors as robots with the highest fitness were achieved when changes in both were allowed.
More, the experiments shed light on another prominent aspect of sensor evolution: economy in the use of resources (precisely sensors in this case).
Indeed, even in the absence of penalty on the number of sensors, the evolutionary algorithm produced robots that used fewer sensors than those available for the task tested.

Like the previous one, many other works in the literature dealing with sensor morphology optimization result in the use of evolutionary algorithms for choosing the range, placement, and type of sensors to be used, as in these works~\cite{auerbach2011evolving,parker2006evolving}, to name a few.

The literature also presents examples of ontogenetic (online) adaptation of sensory systems~\cite{sugiura2010simultaneous}, but again it translates in practical terms into the choice of a possible instantiation of the parameterized sensory system at the design stage, in this specific case involving the number and arrangement of floor sensors.

In~\cite{nurzaman2013active}, a noteworthy work on \emph{in situ} adaptation of sensor morphology is presented.
The study introduces a robotic system that can adapt the end effector of a robot manipulator by extruding thermoplastic adhesive material.
Although the work emphasizes the crucial importance of autonomous adaptation of sensor morphology, the process it uses is expensive, specific to the material used and the goal for which the robot was designed, and therefore not easily generalizable to other application scenarios.
For a complete discussion of sensory morphology adaptation, the reader is referred to the following review paper~\cite{fumiya2016adaptation}.
 
To move from the typical question ``what is the optimal arrangement of your sensors?''~\cite{Eiben2021} to designing robots to cope with uncertain environments through mechanisms that allow them to create \emph{their own sensors}, this paper introduces the concept of ``metasensor.'' 
The ``metasensor'' is a generic architectural component that extends the classical robot model through an additional layer that offers, at the same time, greater sensory capabilities to the robot and a computational model suited to accommodate a process of sensor evolution reminiscent of that taking place in living organisms.

This article is organized as follows. 
Section~\ref{section:metasensor} is devoted to the presentation of the metasensor model and the problems it aims to overcome.
The section~\ref{section:metacybbio} describes the relationship between the metasensor and the research fields of cybernetics and biosemiotics.
Finally, the last section~\ref{section:conclusion} concludes the paper by mentioning some possible real application scenarios and visionary applications involving the employment of the metasensor. 

\section{Metasensor model: control-by-interpretation}\label{section:metasensor}

As can be appreciated from the examples presented in the Introduction, artificial sensor evolution is not an ``all or nothing'' property, but, rather, a characteristic of a system for which a degree of attainment can be defined, which depends on designer's choices and in turn on the possibilities offered by the robot model used. In fact, the evolution of sensors in artificial systems is a process that must be enabled, fostered, explicitly by the designer, not a ``for free'' emerging property provided by natural selection, development and learning processes, as is the case of the biological counterpart.
It then becomes necessary to design an architectural component that can accommodate the artificial evolution of sensors, possibly requiring minimal human intervention and being generic enough to be suitable for various application scenarios.

For these reasons, the novel concept of \textbf{``metasensor''} is introduced here for the first time. 
Figure~\ref{fig:MetasensorSchemaFigure} reports the architectural change to the classical robot sensorimotor loop~\ref{fig:classicalschema} imposed by the introduction of the metasensor~\ref{fig:metasensorschema}.

\newcommand\verticalspacecaption{-1}
\newcommand\verticalspacesubfigure{0}

\begin{figure}[h!]
\captionsetup[subfigure]{justification=centering}	
\centering\begin{subfigure}[b]{0.5\linewidth} 
	\centering\includegraphics[width=1\linewidth]{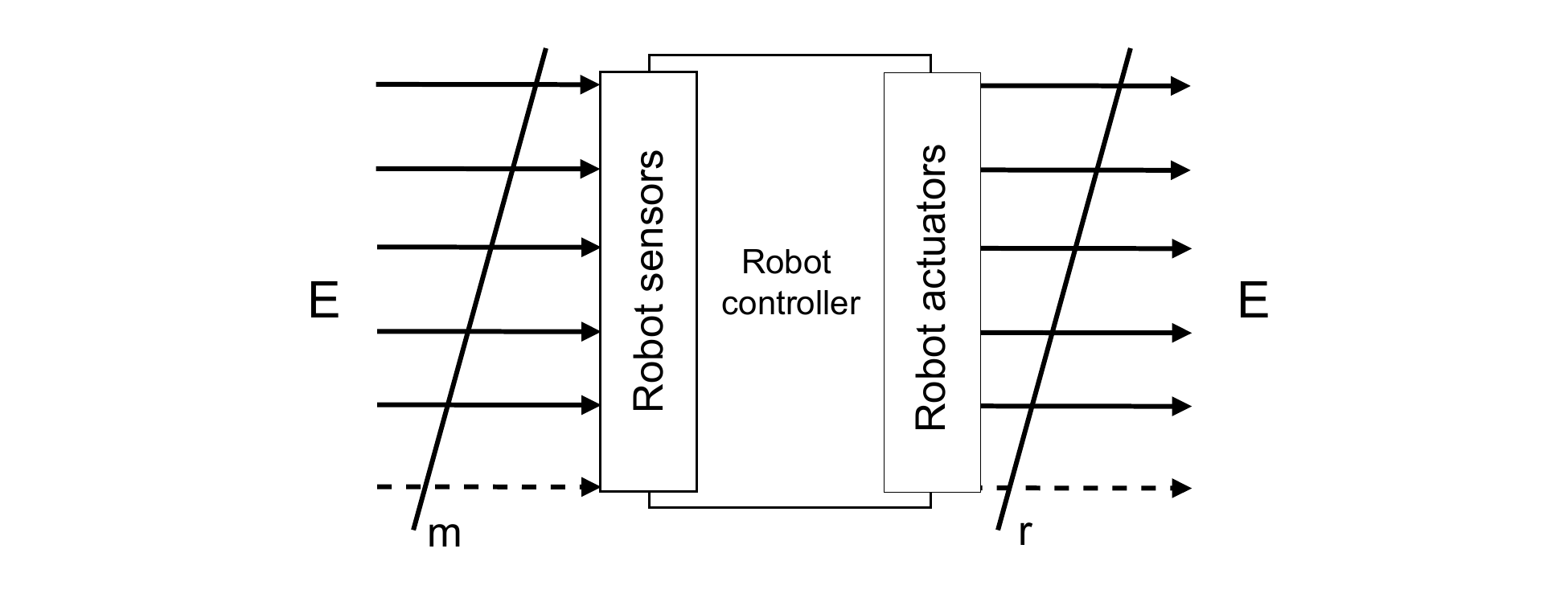} 
 	\vspace{\verticalspacecaption\baselineskip}
	\caption{\label{fig:classicalschema}} 
\end{subfigure}\hfill
\begin{subfigure}[b]{0.5\linewidth} 
	\centering\includegraphics[width=1\linewidth]{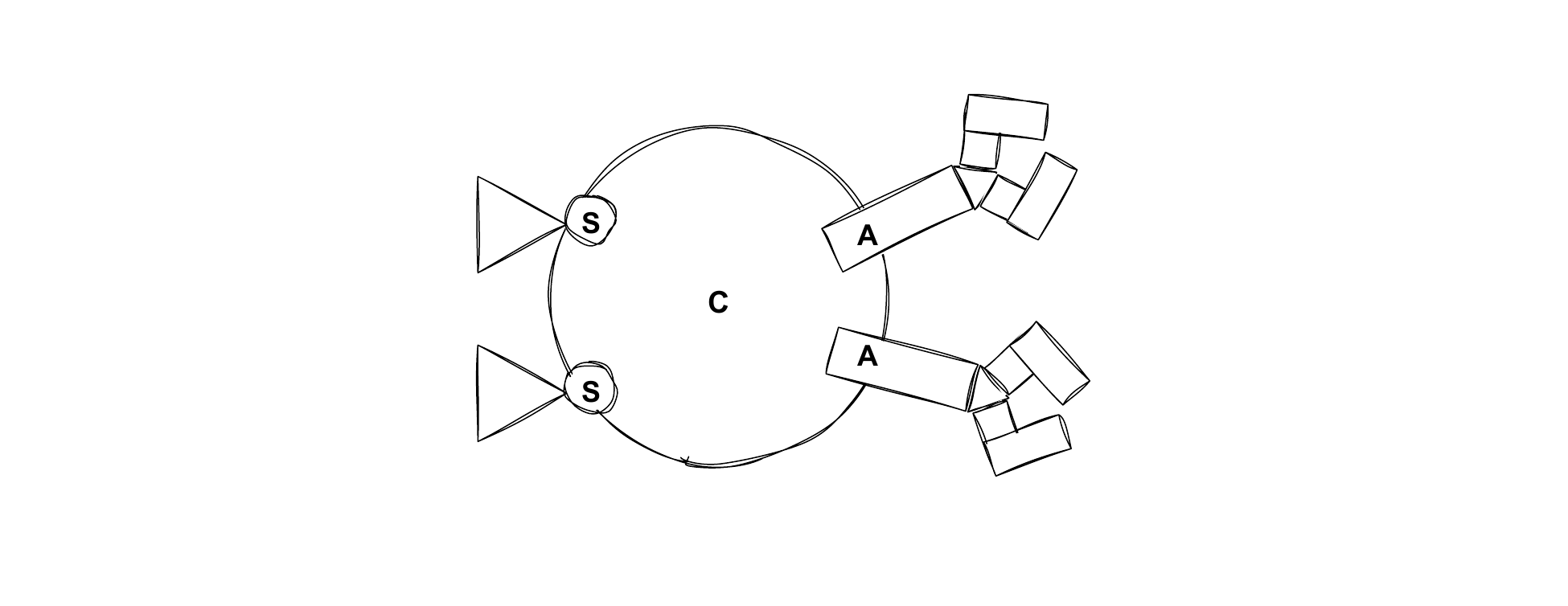} 
	 \vspace{\verticalspacecaption\baselineskip}
	 \caption{\label{fig:classicalrobot}} 
\end{subfigure}\vspace{10pt}

\vspace{\verticalspacesubfigure\baselineskip}

\captionsetup[subfigure]{justification=centering}	
\centering\begin{subfigure}[b]{0.5\linewidth} 
	\centering\includegraphics[width=1\linewidth]{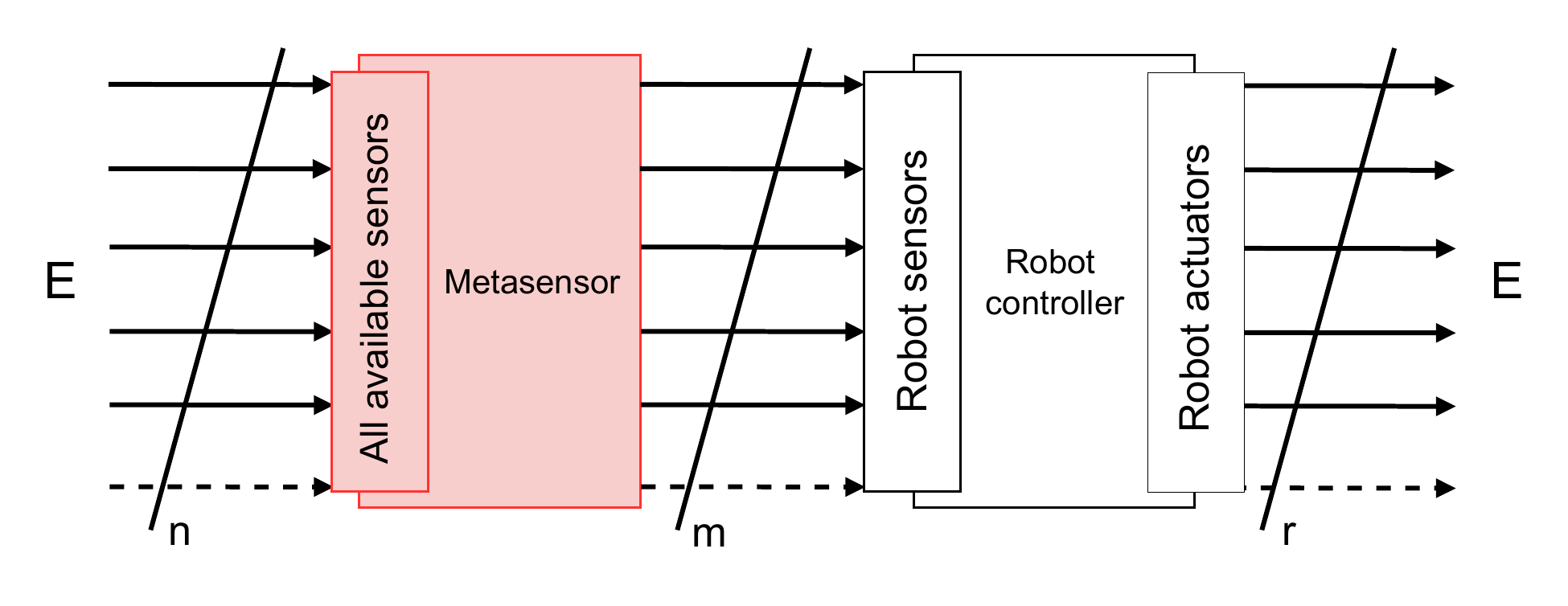} 
 	\vspace{\verticalspacecaption\baselineskip}
	\caption{\label{fig:metasensorschema}} 
\end{subfigure}\hfill
\begin{subfigure}[b]{0.5\linewidth} 
	\centering\includegraphics[width=1\linewidth]{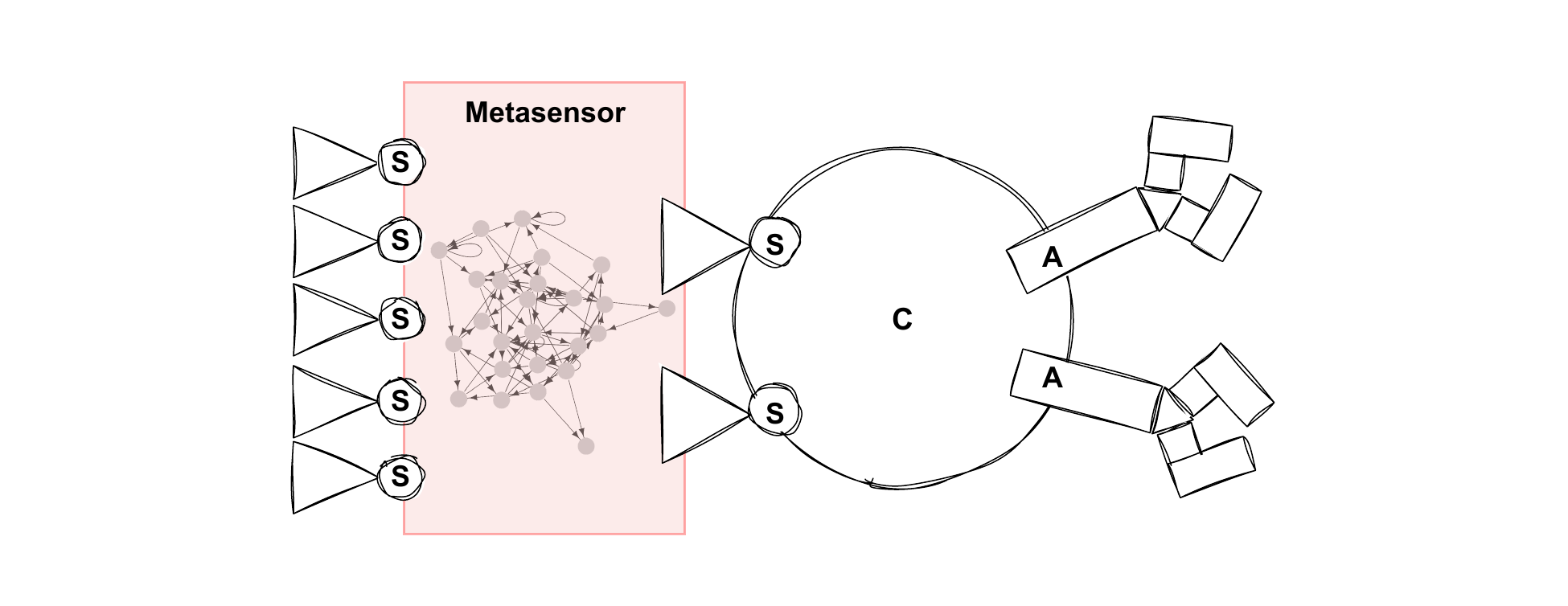} 
	 \vspace{\verticalspacecaption\baselineskip}
	 \caption{\label{fig:robotmetasensor}} 
\end{subfigure}\vspace{10pt}

\caption{
Role of metasensor in the robot sensorimotor loop. \textbf{(a)} Abstract representation of the components of the sensorimotor loop of a robotic agent. \textbf{(b)} Schematic representation of a robot with a sensorimotor loop composed of two sensors (S) two actuators (A) and a control software (C). \textbf{(c)} Diagram representing the role of the \emph{metasensor} in the robot sensorimotor loop: in short, it enhances the sensory capabilities of the robot by increasing the number and type of sensors and offering memory-based processing. \textbf{(d)} Possible instantiation of the metasensor concept with the robot proposed in \textbf{(b)}: an example of the internal system that composes the metasensor processing unit in the form of a recurrent network is illustrated.
} 
 \label{fig:MetasensorSchemaFigure}
\end{figure}

From the schema reported in figure~\ref{fig:metasensorschema} it can be noted that the metasensor is composed of two parts: \emph{(i)} a hardware interface responsible for amplifying the robot \textbf{sensing} through the introduction of a series of physical sensors capable of capturing signals coming from different phenomena present in the environment; \emph{(ii)} a software computational module responsible for the \textbf{perception} phase which allows higher order processing and therefore the integration, interpretation and organization of the captured sensory information. 

Referring to figure~\ref{fig:metasensorschema}, the hardware component extends the vector of robot input signals $\mathbf{I} = \begin{bmatrix} I_1, I_2, \dots,  I_m \end{bmatrix}^T$ from $m$ to $n$ dimensions, with in general $n \ge m$~\footnote{The $n=m$ case represents an interesting case only if the sensory modalities are different from those the robotic model used is endowed with.}.
It is also possible that not all of the $m$ sensors of the robot are controlled by the metasensor. In this case, some sensors will continue to capture signals from the surrounding environment.
In principle, we could state that the more generic the sensors are (i.e., with input values in a wide range and possibly without any kind of signal preprocessing), the more freely the mechanism of sensor evolution that takes place in the metasensor can explore new correlations between signals and actions, and thus create new signs and, in turn, new meanings useful for successfully attaining any given task.
Ideally, a suitable generic sensor for this type of application would be broadband sensors capable of capturing electromagnetic waves over a wide range of frequencies, such as the one reported in~\cite{liu2022broadband}, a photonic electric field sensor capable of capturing a frequency range from 10 MHz to 26.5 GHz.
Clearly, there must be compatibility between the metasensor's output interface and the robot's input interface: in other words, the metasensor must respect the physical constraints imposed by the robot by producing output signals that can be interpreted by its sensory modalities.
For example, a metasensor connected to a robot equipped only with light sensors will have to emit light signals in order to control it.

On the software side, the metasensor autonomously performs a \textbf{model-free} search in the space of ``perceptual processes'' with the goal of finding the optimal interpretation of its input signals to feed to the robot to make it perform the given task.
To this end, although in principle there are no constraints on the model that can reify the metasensor at a practical level, subsymbolic models are an ideal representation of it, as they operate at a lower level of abstraction than symbolic models and can leave full freedom to evolution to achieve the best configuration for the particular scenario at hand.
The metasensor dynamically changes---through evolution/adaptation/self-organizing mechanisms---the way it processes the input signals just captured by the hardware interface.
The outputs thus computed will in turn become the robot's inputs and act on it by perturbing its dynamics.
The meaning (semantics) attributed to the input signal by the metasensor will be contingent on the specific environment, task, robot model, and the specific affordances~\cite{gibson1966senses,roli2024cybernetic,roli2020emergence} that become available over time.
The type of control exerted by the metasensor on the robot can be referred to as ``control-by-interpretation'': the metasensor tries to find the appropriate interpretation of the input signals able to exploit the bouquet of dynamics that the robot can exhibit to steer it toward the desired behavior. 
It is precisely the ability to perform a model-free search in the space of perceptual processes that endows the entire robot-metasensor system with a \emph{semantic adaptivity} and so \emph{the ability to evolve its own sensors} to cope with the contingent situation.
The metasensor acts from a semiotics perspective as a meaning-making subsystem.
Indeed, sensor evolution allows the robot to successfully operate in unpredicatable and dynamics environments: metasensor can change---as a function of a potentially changing over time objective function---to deal with the contingencies as they arise.

Unlike the classical robotic scheme shown in the figure~\ref{fig:classicalschema}, the signals that perturbs the robot are now influenced not only by the robot's environment and actions, but also by the internal dynamics (the process of meaning creation) of the metasensor.
This can be formalized as follows:
\[
	\mathbf{O}_M(t + 1) \equiv \mathbf{I}_R(t + 1)  = f(\mathbf{E}(t), \mathbf{U}_M(t),  \mathbf{A}_R(t), t )	
\]
where $\mathbf{O}_M = (o_1, o_2, \dots, o_m)$ and $\mathbf{U}_M$ refer to the output vector and the state vector of the metasensor, respectively; while $\mathbf{E}$ represents the state of the environment.
Additionally, $\mathbf{I}_R$ and $ \mathbf{A}_R= (a_1, a_2, \dots, a_r)$ are the inputs controlled by the metasensor and the action vector of the robot.
Since the metasensor can retain memory of its past, the state vector  $\mathbf{U}_M$ can depend on its previous state, going thus beyond the concept of a dynamical multiplexer:
\[
	 \mathbf{U}_M(t+1)  = g(\mathbf{E}(t), \mathbf{U}_M(t),\mathbf{U}_M(t-1), \ldots, \mathbf{U}_M(0), \mathbf{A}_R(t), t )	
\]
The dependence of the functions $f$ and $g$ on time accurately reflects the possibility of a contingent evolution of the metasensor.
In summary, the dynamics of the metasensor, together with the environment and the robot's own past actions, determine the sensory inputs of the controlled robot.
This will in turn affect the robot's internal state and its subsequent actions, thereby closing the causal feedback structure---between environment, metasensor, and robot---that drives the robot's observable emergent behavior.
From the perspective of information theory, the metasensor acts on signal interpretation to reduce uncertainty about the input signal itself.

The sensor evolution mediated by the metasensor is a complementary approach to robot control software design.
However, it offers the advantage of being able to operate in application scenarios where changing the control software is problematic for various reasons.
The advantages become apparent when it is impossible to change the controller of an already operating robot in the real world either because it is deemed too expensive or because the controller's internal model is inaccessible.
In those cases the metasensor represents a viable control mechanism for modifying the robot's behavior while leaving its controls software unchanged, even and especially in online scenarios, i.e., taking place during the robot's lifetime~\cite{baldini2023performance,braccini2022artificial}.

\section{Between Cybernetics and (Bio)Semiotics} \label{section:metacybbio}

From the argumentation provided in the previous sections, it is evident how the action of the metasensor is aimed at controlling a robotic agent by allowing it to express different behaviour from that for which it was designed, and thus how this component acts as a cybernetic component.
Indeed, it plays a fundamental role in determining ``what the robot does''.
In this sense, we can frame the metasensor as a generalization of Arkin's Perceptual Schema~\cite{arkin1987motor}, Selfridge's Pandemonium architecture~\cite{selfridge1988pandemonium} and, in general, the sensory input layer of many other control schemas that fall under the cybernetic discipline.

However, if we ask ``what is'' the metasensor, it is inevitable to state that it is the component in which the phenomena of signification reside and occur, and thus its semiotic role in the robot-environment relationship becomes evident.
Cariani in his work~\cite{cariani1992some,cariani1993evolve,cariani1990adaptive,cariani1989design}
 had already stressed the importance of the creation of meaning and appropriate \emph{relevance criteria} in artificial agents and how they foster the creation of autonomous agents, in the true sense of the word.
In part, while not defining a model, Cariani himself had historically anticipated the role and importance of the metasensor, speaking of ``open-ended neural networks'' for creating ``semantically adaptive artificial devices''.

It is, however, with the conceptual decoupling of metasensor and robot controller, first proposed in this work, and with the following recently published work~\cite{braccini2022artificial, braccini2022criticality, baldini2023performance}, that the foundations have been laid for achieving the evolution of sensors in artificial contexts.
These works indeed paved the way for the realization of the metasensor layer from a practical point of view, as they present the possibility of implementing a non-destructive adaptation process acting on a raw substrate, e.g., Boolean networks and nanowire networks~\cite{milano2020brain}, in robotic contexts, in order to achieve different behaviors.
By acting only on the sensor-node and node-actuator couplings without altering the structure of these models, the authors showed that robots can show interesting behaviors.
Thus, the same unchanged network, and especially the bouquet of dynamics it can express, can be reused to make the robot controlled by it perform another, novel, behavior.

The same process is at the basis of metasensor functioning, but what is gradually crafted by the adaptation process is not the control software, but the interpretation of the input signals that will later feed into the robot's control software.
The evolution in the case of metasensor works in the space of symbols received by the input sources and on the already processed chunk of symbols, generating, depending on the specific circumstances, information processing processes useful for obtaining the desired behavior.
This process can lead to the creation of new internal symbols and new interpretations of old symbols, thus extending the syntactic and semantic vocabulary of the metasensor-robot system~\cite{roli2024cybernetic}.
In short, we can say that the evolution of sensors in the metasensor assumes the form of the search in the space of \emph{epistemic functions}~\cite{pattee2013epistemic}.

\section{Conclusion}\label{section:conclusion}

This paper presented a novel architectural component capable of controlling and, at the same time, making semantically adaptive an already developed---and operating---robotic agent.
The metasensor exploits the information provided by its sensory input layer and, more importantly, the inherent plasticity offered by sub-symbolic computational models in concert with online adaptation mechanisms, to modify the robot's behavior, leaving its hardware and software equipment unaltered.
 
Future work will be devoted to developing application scenarios to test and appreciate its potential in robotics.
Its properties meet the needs and constraints imposed by online robot adaptation.
Thus, case studies can be envisioned in which, during the robot lifetime, a robot is assigned a task for which it lacks the sensory modalities necessary for its successful accomplishment.
So, the metasensor has to adapt its perception---through a reorganization of the set of information processing dynamics---to provide appropriate processing of the new information to accomplish the new task, while syntactically respecting the constraints expressed by the robot's sensory interface.
This also implicitly addresses the issues of optimization and reuse of the available resources, minimizing costs since it avoids a complete redesign of the robotic agent.

Visionary applications of the metasensor involve robotic applications in hostile environments, where groups of simple (micro)robots must complete a given mission. 
Here individual (or groups of) robots may (i) need to change over time the subtask they face and (ii) be subject to possible sources of information not foreseen in advance by the designer, or a combination of the two.
In any case, the metasensor provides the robot with the ability to flexibly adapt its behavior on the fly.

Lastly, since it can be implemented on physical hardware, the metasensor offers the possibility to investigate whether it is possible to overcome in robotic applications the concept of combinatorial novelty, i.e. the same emergent phenomenon that evolution triggers in Thompson's evolvable hardware.

\section*{Acknowledgments}
I gratefully acknowledge the useful and inspiring discussions with Andrea Roli and Paolo Baldini (Universit{\`a} di Bologna).

\bibliographystyle{ieeetr} 
\bibliography{main}

\end{document}